%% file: root.tex
\newcommand{\CLASSINPUTinnersidemargin}{19.1mm}
\newcommand{\CLASSINPUTtoptextmargin}{19.1mm}
\newcommand{\CLASSINPUTbottomtextmargin}{19.1mm}
\documentclass[conference]{IEEEtran}
\IEEEoverridecommandlockouts
\usepackage{cite}
\usepackage{amsmath,amssymb,amsfonts}
\usepackage{algorithmic}
\usepackage{graphicx}
\usepackage{textcomp}
\usepackage{xcolor}
\usepackage[dvipsnames]{xcolor}
\usepackage{multirow}
\usepackage{svg}
\usepackage{caption}
\usepackage{subcaption}
\usepackage{amsmath}
\usepackage{makecell}
\usepackage{float}
\usepackage{rotating}
\usepackage{hyperref}
\usepackage{booktabs}
\def\BibTeX{{\rm B\kern-.05em{\sc i\kern-.025em b}\kern-.08em
    T\kern-.1667em\lower.7ex\hbox{E}\kern-.125emX}}

\newcommand{\checkedsquare}{\makebox[0pt][l]{$\square$}\raisebox{.15ex}{\hspace{0.1em}$\checkmark$}}
\newcommand{\yesbox}{\checkedsquare}
\newcommand{\nobox}{\square}

\begin{document}

\title{\vspace*{18pt}Context-Aware Model-Based Reinforcement Learning for Autonomous Racing
\thanks{\(^1\)Emran Yasser Moustafa and Ivana Dusparic are with the School of Computer Science and Statistics, Trinity College Dublin, Ireland (email: moustafe, duspari@tcd.ie)}
}

\author{
\IEEEauthorblockN{Emran Yasser Moustafa\(^1\) and Ivana Dusparic\(^1\)}
}

\maketitle

\input{abstract}

\input{intro}

\input{relatedwork}


\input{cmask}

\input{experiments}

\input{results}

\input{conclusion}

\input{acknowledgment}

\input{bibliography}

\end{document}

%% file: abstract.tex
\begin{abstract}
Autonomous vehicles have shown promising potential to be a groundbreaking technology for improving the safety of road users. For these vehicles, as well as many other safety-critical robotic technologies, to be deployed in real-world applications, we require algorithms that can generalize well to unseen scenarios and data. Model-based reinforcement learning algorithms (MBRL) have demonstrated state-of-the-art performance and data efficiency across a diverse set of domains. However, these algorithms have also shown susceptibility to changes in the environment and its transition dynamics. 

In this work, we explore the performance and generalization capabilities of MBRL algorithms for autonomous driving, specifically in the simulated autonomous racing environment, Roboracer (formerly F1Tenth). We frame the head-to-head racing task as a learning problem using contextual Markov decision processes and parameterize the driving behavior of the adversaries using the context of the episode, thereby also parameterizing the transition and reward dynamics. We benchmark the behavior of MBRL algorithms in this environment and propose a novel context-aware extension of the existing literature, cMask. We demonstrate that context-aware MBRL algorithms generalize better to out-of-distribution adversary behaviors relative to context-free approaches. We also demonstrate that cMask displays strong generalization capabilities, as well as further performance improvement relative to other context-aware MBRL approaches when racing against adversaries with in-distribution behaviors.


\end{abstract}


%% file: intro.tex
\section{Introduction}

Reinforcement learning (RL) is increasingly being used as a method of autonomous control in robotics. In applications such as autonomous racing, RL-powered agents have achieved super-human levels of performance \cite{fuchs2021super, wurman2022outracing, vasco2024super}. Much of this success has been achieved by model-free RL (MFRL) methods, such as extensions of the popular SAC algorithm \cite{haarnoja2018soft}. However, in a diverse set of domains, model-based RL (MBRL) algorithms have demonstrated performance comparable to their MFRL counterparts \cite{hafner2019dream, hafner2020mastering, hafner2023mastering}. MBRL algorithms work by learning to maximize the rewards received by the agent, as well as learning a model of the environment and its changes over time. In non-stationary environments, however, the transition dynamics of the underlying MDP may be difficult to predict, potentially leading to failure \cite{kirk2023survey, prasanna2024dreaming}.

These characteristics of MBRL algorithms may make them unsuitable for many safety-critical robotics applications, such as autonomous driving. Interactions with road users are varied and highly unpredictable, especially in urban driving scenarios \cite{boekema2024multi, boekema2025road}. If we consider the control of an autonomous vehicle as an RL problem, the actions taken by road users may be seen as introducing non-stationarity to the environment and make modeling the transition dynamics a more difficult task. In order for MBRL algorithms to be used in such real-world control applications, new methods must be proposed that generate policies which are more capable of robustly generalizing to unseen changes in the environment such as unseen behaviors by other road users.

In this work, we evaluate the performance and generalization capabilities of MBRL algorithms in the autonomous racing environment, Roboracer (formerly F1Tenth). We do so by framing the head-to-head racing task as a reinforcement learning problem modeled with contextual Markov decision processes. We parameterize the driving behavior of adversaries in the environment using the context, thereby also parameterizing the transition and reward dynamics. In our experiments, we train MBRL agents under a restricted set of contexts, however, evaluate these agents under a wider set of contexts. This allows one to understand each algorithms ability to generalize to unseen road agent behavior. It is in this setting that we propose an extension of the context-aware Dreamer algorithm proposed by Prasanna et al. \cite{prasanna2024dreaming}. Our method, cMask, is designed to suit environments in which the context may not always be relevant to planning actions or predicting future states, head-to-head autonomous racing for example. The algorithm uses a SAC network to remap the context values of the episode before using them in the world model. We demonstrate that our method displays strong generalization capabilities relative to other MBRL methods, as well as performance improvements when racing against adversaries with in-distribution behaviors. We also demonstrate that the use of context-aware MBRL approaches in our environment results in policies that both generalize better and are safer in agent-to-agent interactions relative to context-free MBRL approaches.


Our contributions can be summarized as follows.
\begin{itemize}
    \item We present a framework for the head-to-head racing task in the Roboracer environment using contextual Markov decision processes. 
    \item We propose cMask, a context-aware MBRL algorithm, which demonstrates generalizable driving behaviors across out-of-distribution contexts and outperforms alternative MBRL methods in select autonomous racing settings.  
    \item We compare the performance and generalization capabilities of MBRL algorithms when racing against adversaries that display in- and out-of-distribution driving behavior.
\end{itemize}

%% file: relatedwork.tex
\section{Related Work}
In this section, we will discuss the body of literature that exists around the two major themes of this work; model-based reinforcement learning, and the domain of autonomous racing.

\subsection{Model-Based Reinforcement Learning}
MBRL algorithms, such as the state-of-the-art DreamerV3 algorithm, have demonstrated the ability to learn complex control policies in partially observable environments with a superior sample efficiency relative to MFRL algorithms \cite{hafner2019dream, hafner2020mastering, hafner2023mastering}. Dreamer learns a compact representation of the environment by reconstructing the input observation from the latent state of a recurrent state-space model (RSSM) \cite{hafner2019learning}. The RSSM model is used to predict future states and rewards given a series of actions are taken, thus improving decision-making capabilities.

However, the performance of MBRL algorithms appears to be sensitive to small changes in the environments transition dynamics \cite{kirk2023survey}, especially when these dynamics are out-of-distribution \cite{prasanna2024dreaming}. For these algorithms to be deployed in real-world robotics, especially safety-critical applications such as autonomous driving, they must be capable of generalizing to unseen dynamics in a robust manner. In this work, we are interested in exploring the how MBRL algorithms generalize across non-stationary environments with differing transition dynamics. Particularly, how these algorithms adapt to unseen adversary behaviors in non-collaborative multi-actor games. 


In order to study this, many works have parameterized features of the environment using a \textit{context} vector. The context may then be sampled so as to induce in-distribution and out-of-distribution worlds to evaluate the RL agent in. In the domain of robotics, works have used an RL agent to learn the control of an aircraft with the context defining the weight of a payload  \cite{belkhale2021model} or the effects of wind \cite{10.1145/1569901.1569922}. Benjamins et al. \cite{benjamins2023contextualizecasecontext} proposed the CARL benchmark, a set of tasks in which the context defines physical parameters of the environment, in order to compare context-aware RL algorithms.


Prasanna et al. \cite{prasanna2024dreaming} proposed a context-aware extension of the RSSM structure, coined cRSSM, which enabled the DreamerV3 algorithm to generalize better relative to alternative approaches when observing out-of-distribution transition dynamics in CARL tasks. Our proposed approach extends this work by further improving performance in out-of-distribution settings, particularly in non-stationary environments.  



\subsection{Autonomous Racing}
Autonomous racing has been viewed as a useful proxy problem to understanding challenges in autonomous driving, especially in the topics of planning and control \cite{betz2019can, betz2022autonomous}. Works in the field of autonomous racing explore how autonomous driving systems operate under high speed and high uncertainty. AVs in this setting must navigate quickly and safely in a constantly changing environment. Roboracer \cite{o2020f1tenth} is a popular testbed and platform for the development of safe autonomous racing algorithms. Vehicles in this domain consist of 1:10-scale remote-controlled cars that have been fit with a 2D LiDAR scanner and onboard compute. Roboracer host virtual and real-world competitions which consist of single-agent and head-to-head racing events. 

Roboracer algorithms can be divided into two categories; rules-based approaches and learning-based approaches \cite{evans2024unifying}. 


\subsubsection{Rules-Based Approaches}
Rules-based approaches rely on explicit algorithmic logic to determine the speed and steering actions that the vehicle should perform. Classic rules-based approaches to Roboracer use a combination of optimal trajectory calculation and a low-level controller for path-following \cite{evans2024unifying}. Heilmeier et al. \cite{heilmeier2020minimum} proposed a framework to generate an optimal global trajectory using quadratic programming under minimum-time and minimum-curvature constraints. Regarding path-following algorithms, pure pursuit has been used extensively in the field \cite{coulter1992implementation, o2020tunercar, becker2023model}. Pure pursuit is a rules-based algorithm that uses a geometric model to determine the steering angle that will keep the vehicle on a given trajectory. 

\subsubsection{Learning-Based Approaches}
In contrast to rules-based approaches, learning-based approaches use data-driven models, often trained via machine learning, to select speed and steering actions rather than algorithmic logic. Deep learning has been extensively explored in Roboracer \cite{evans2024unifying}. Many approaches use handcrafted reward functions to train agents to show desired behaviors, such as driving at high speeds \cite{bosello2022train}, following a race line trajectory \cite{ghignone2023tc, evans2023high}, or overtaking \cite{trumpp2024racemop}. 

Brunnbauer et al. \cite{brunnbauer2022latent} proposed a Dreamer adaptation for single-agent autonomous racing. Their agent demonstrated obstacle avoidance behavior. However, further work has shown that this method achieves slower lap times compared to MFRL methods \cite{zhang2022residual, evans2023high}. These works were also conducted in a single-agent racing scenarios only, i.e. environments in which there is a single vehicle on the race track controlled by the RL agent. To the best our knowledge, our work is the first to explore the use of MBRL in Roboracer for head-to-head racing, i.e. environments in which there are multiple competing vehicles on the race track, however, only a single vehicle is controllable by the RL agent. Model-based approaches to head-to-head racing have been explored in other autonomous racing environments, demonstrating the Dreamer algorithm's ability to learn complex driving maneuvers such as blocking, overtaking and evading other drivers \cite{schwarting2021deep}.


%% file: cmask.tex
\section{Methodology}

\subsection{RL Preliminary}\label{sec:marl-prelim}

In this work, the learning problem is framed as a single-agent RL problem in a partially observable, context-defined environment. We model interactions in this setting using an adaption of a contextual MDP (cMDP) \cite{ghosh2021generalization}, in which the transition and reward dynamics of the episode are conditioned on a context variable $c \in \mathcal{C}$. 

\subsection{cMask Algorithm}
The cMask algorithm is a context-aware extension of the DreamerV3 algorithm \cite{hafner2019dream,hafner2020mastering,hafner2023mastering}. The algorithm substitutes the RSSM structure with a cRSSM structure \cite{prasanna2024dreaming}, however, uses a SAC network to predict an attenuating mask \(m_t\) that is multiplied element-wise with the context before it is concatenated with the model state. The mask size matches that of the context vector and its values are restricted to the range \([0, 1]\). 

\begin{align*}
     c^{m} \sim m_t * c, \quad m_t \in \{0, 1\}  \\ 
     \label{eq:mask}
\end{align*}

The design of the cMask algorithm is motivated by the insight that in many tasks the effects of context, or the elements in the environment that affect the transition and reward dynamics, are not always relevant to RL agent's policy in selecting the optimal action or predicting future states. We refer to environments of this kind as non-stationary as, from the perspective of the RL agent, the transition dynamics may appear not to be constant throughout the episode or only partially dependent on the context. This is in contrast to the original cRSSM paper where the dynamics of the environment in the selected task appear to be parameterized by the context at every time step. The naive use of the cRSSM structure in non-stationary environments may lead to an unnecessarily large state space in which the context may act as a distractor and impair policy robustness. Our intuition behind the design of cMask is that the SAC network may independently learn to selectively remap the context vector for when it is relevant and when it is not relevant, and thereby improve policy performance in non-stationary environments.

Regarding the mask network, the actor predicts the attenuating mask conditioned on the observation \(o_t\), while the critic predicts the distribution of the expected discounted future rewards \(G_t\) conditioned on the observation \(o_t\) and the predicted attenuating mask \(m_t\). The SAC model is trained using the extrinsic reward signal from the environment, \(r_t\).

\begin{align*}
    \text{Mask Actor:} & \quad m_t \sim \pi_\phi(m_t | o_t) \\
    \text{Mask Critic:} & \quad v_\phi(G_t | o_t, m_t) \approx \mathrm{E}_{\pi_\phi(\cdot| o_t, m_t)}[\Sigma_{\tau=0}^{H-t}r_{t+\tau}] \\
\end{align*}

The world model equations are defined in a similar manner as the original cRSSM paper \cite{prasanna2024dreaming}, however, the context vector \(c\) is replaced with a masked context \(c^m\). 


\subsection{cMask in Roboracer}
To apply the cMask algorithm in the Roboracer environment, we frame the environment and its interaction with the algorithm in terms of standard RL constructs. The following is a description of states, observations, actions, the reward function, and the definition of an episode used in this work. 

\textbf{State:} Vehicle states are represented using a 7-dimensional state vector containing the agents \(x\) and \(y\) position, longitudinal velocity \(v\), orientation (yaw) angle \(\theta\), yaw rate \(\dot{\theta}\), steering angle \(\delta\), and slip angle \(\beta\). All state features are relative to the global frame, which is located at the trainable agent's starting position and is oriented with the trainable agent's initial heading (zero yaw). Vehicle dynamics are modeled and updated using a single-track bicycle model \cite{althoff2017commonroad}.

\textbf{Observation:} The trainable agent uses an observation rather than a state to determine the appropriate action. The observation consists of a 108-element vector representing a 2D LiDAR scan of the environment in a 180\(^\circ\) FOV in front of the vehicle. Each element in the vector, representing a single beam in the scan, is modeled as a ray cast outward from the center point of the vehicle at an angle relative to the vehicle's heading. The magnitude of the element indicates the distance the ray traveled before intersecting with an obstruction, i.e. the track boundaries or another vehicle. The value of each beam \(d\) is scaled to the range \([0, 1]\) relative to a configurable maximum beam length value.

\textbf{Action:} The action space for each agent consists of a tuple of two continuous scalar outputs, representing the speed and steering angle of the vehicle. Both values are in the range \([-1, 1]\). The steering angle is scaled according to the maximum steering angle \(\theta_{max}\) to a range \([-\theta_{max}, \theta_{max}]\). The negative and positive values of \(\hat\theta_t\) denote requests to turn left and right, respectively. The velocity \(\hat{v}_t\) is also scaled to a range \([0, v_{max}]\) where \(v_{max}\) denotes the maximum velocity of the vehicle.

\textbf{Reward:} Following Jaritz et al. \cite{jaritz2018end}, each agent receives a cross-track and heading reward signal. The reward function is shown in Eq. \ref{eq_cth}, where \(v_t\) is the vehicle velocity, \(v_{max}\) is the maximum vehicle velocity,  \(\phi\) is the difference in heading angle between the vehicle and the road and \(d_c\) is the minimum distance between the vehicle and the center line of the track. Agents receive an additional reward of \(-1\) for crashing and a reward of \(+1\) is when an agent completes a lap of the track. Agents also receive a reward of \(+1\) each time they overtake another agent, and a reward of \(-1\) each time they have been overtaken.

\begin{equation}
        r_t = \frac{v_t}{v_{max}}\cos(\phi) - d_c
        \label{eq_cth}
\end{equation}

\textbf{Episodes:} Episodes consist of a single lap of the race environment. The position of the starting/finishing line is set to the initial position of the trainable agent. Episodes conclude when the trainable agent crosses the finish line or is involved in a collision with an adversarial agent or the track boundaries. Episodes begin with all agents in a straight line with the trainable agent always being located at the back of the line. Spacing between the agents is dependent on the track layout selected.

%% file: experiments.tex
\section{Experiments}

We evaluate our approach using the open-source Roboracer Gym simulator presented by O'Kelly et al. \cite{o2020f1tenth}. Gaussian noise with a standard deviation of 0.01 is added to each beam in the LiDAR scan before being passed to the corresponding agent. No noise is added to the actions of agents before being executed. The dynamics of the environment are updated at 100 Hz, however each agent interacts with the environment at 10 Hz. All agents receive their observation and perform planning in the same frame. 


\subsection{Adversaries}
Adversaries in all experiments use a combination of a global optimal trajectory and a local path-following algorithm to navigate the track. An optimal trajectory is calculated offline using the method proposed by Heilmeier et al \cite{heilmeier2020minimum}. The pure pursuit algorithm \cite{coulter1992implementation} is then used to select the steering action that will keep the vehicle on the trajectory. The adversary selects a speed action that corresponds to the speed associated with the lookahead point used by the pure pursuit algorithm. In order to encourage more interactions between the trainable agent and adversaries in the environment, the speed action selected by the adversary is reduced by 70\%. 

\subsection{Use of Context}
\begin{figure*}
    \vspace{2pt}
    \centering
    \includegraphics[width=0.85\linewidth]{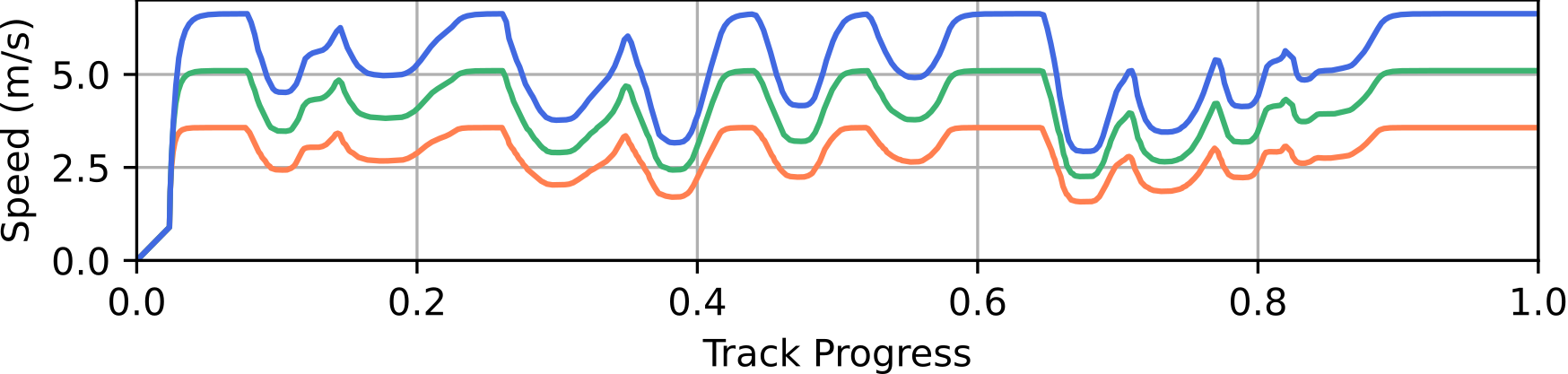}
    \caption{Velocity profiles of adversaries in ESP track subject to different velocity contexts. All adversaries in this figure have a steering context \(c_\theta = +0.0\). The \textcolor{RedOrange}{orange}, \textcolor{Green}{green} and \textcolor{RoyalBlue}{blue} lines correspond to the velocity profiles of adversaries with a velocity context equal to \textcolor{RedOrange}{\(c_v = -0.3\)}, \textcolor{Green}{\(c_v= +0.0\)} and \textcolor{RoyalBlue}{\(c_v = +0.3\)}, respectively.}
    \label{fig:pp-speed}
\end{figure*}

\begin{figure}
    \centering
    \includegraphics[width=0.95\linewidth]{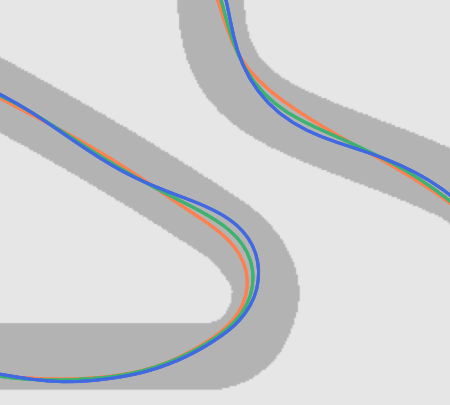}
    \captionsetup{width=\linewidth}
    \caption{Race lines of adversaries in ESP track subject to different steering contexts. All adversaries in this figure have a velocity context \(c_v = +0.0\). Race lines colored in \textcolor{RedOrange}{orange}, \textcolor{Green}{green} and \textcolor{RoyalBlue}{blue} correspond to the routes taken by adversaries with a steering context equal to \textcolor{RedOrange}{\(c_\theta = -0.3\)}, \textcolor{Green}{\(c_\theta = +0.0\)} and \textcolor{RoyalBlue}{\(c_\theta = +0.3\)}, respectively.}
    \label{fig:pp-steering}
\end{figure}

A two-element context vector \(c=[c_v, c_{\theta}]\) is initialized at the beginning of each episode and remains unchanged throughout the duration of the episode. The context is used to parameterize the driving behavior (speed and steering) of all adversaries in the episode. After each adversary selects a speed action and the value is attenuated by 70\%, the resulting value is scaled by a factor equal to \(1 + \lambda_vc_v\). Therefore, the speed coefficient \(c_v\) controls the magnitude of the speed profile of each adversary throughout the race. Similarly, the lookahead distance used in the pure pursuit algorithm is scaled by a factor equal to \(1+\lambda_{\theta}c_{\theta}\). As the lookahead distance controls how "sharply" the agent steers \cite{coulter1992implementation}, the steering coefficient controls the magnitude of the steering profile of each adversary throughout the race. \(\lambda_v\) and \(\lambda_\theta\) are scalar hyperparameters for the speed and steering coefficients, respectively, and are selected based on the track layout. Figs. \ref{fig:pp-speed}-\ref{fig:pp-steering} display the effects of context on the speed and steering profile of the adversaries throughout the race.

\subsection{Baselines}
We compare the performance of our proposed approach to that of three baselines. These baselines are as follows:

\textbf{SAC} An implementation of the model-free SAC algorithm as presented by \cite{haarnoja2018soft}. As SAC does not use a recurrent state, the observation from the previous and current frame are stacked before being used as input to the algorithm.

\textbf{DreamerV3} An implementation of the DreamerV3 algorithm as presented by \cite{hafner2023mastering}. Following previous implementations of the Dreamer algorithm in the Roboracer environment \cite{brunnbauer2022latent}, our implementation uses MLP to encode the input observations rather than a CNN.

\textbf{cRSSM} We extend our implementation of the DreamerV3 agent to use a cRSSM structure \cite{prasanna2024dreaming} rather than an RSSM structure \cite{hafner2019learning}.  In the remainder of the paper, we will use the terms cRSSM agent when referring to the DreamerV3-based agent that uses the cRSSM structure.


\subsection{Hyperparameters}
The neural networks used in the SAC agent were composed of two fully connected layers with 100 neurons each. Each layer used a \textit{ReLU} activation function, and a \textit{tanh} activation function was used to map the output to the range \([-1, 1]\).The networks were trained using a batch size of \(100\), a learning rate of \(3\times10^{-5}\), and a discount factor of \(0.85\). 

The DreamerV3, cRSSM and cMask agents used the same hyperparameters. In the world model component of each algorithm, the number of GRU units is equal to 256. The encoder-decoder components used a three-layer MLP with 256 neurons per layer. A batch size of 16, batch length of 64, and learning rate of \(3\times10^{-5}\) was used. An imagination horizon of length 10 was also used. All other hyperparameters were chosen from the \textit{small} variant of original DreamerV3 network \cite{hafner2023mastering}.

\subsection{Experiments}
The performance of each algorithm is evaluated in a race scenario against a single adversary and against three adversaries. At the beginning of training, both elements in the context vector are initialized independently from the range \([-0.15, +0.15]\). During evaluation, each element in the context vector is selected iteratively in a grid-search manner from a range \([-0.3, 0.3]\) using a step size of \(0.1\), totaling 49 unique context configurations. As the context controls the driving behavior of the adversaries in the episode, the trainable agent's performance can be evaluated against adversaries that display in- and out-of-distribution behaviors. 50 test laps are conducted for each unique context configuration. 

\begin{figure}[H]
    \centering
    \includegraphics[width=0.4\linewidth]{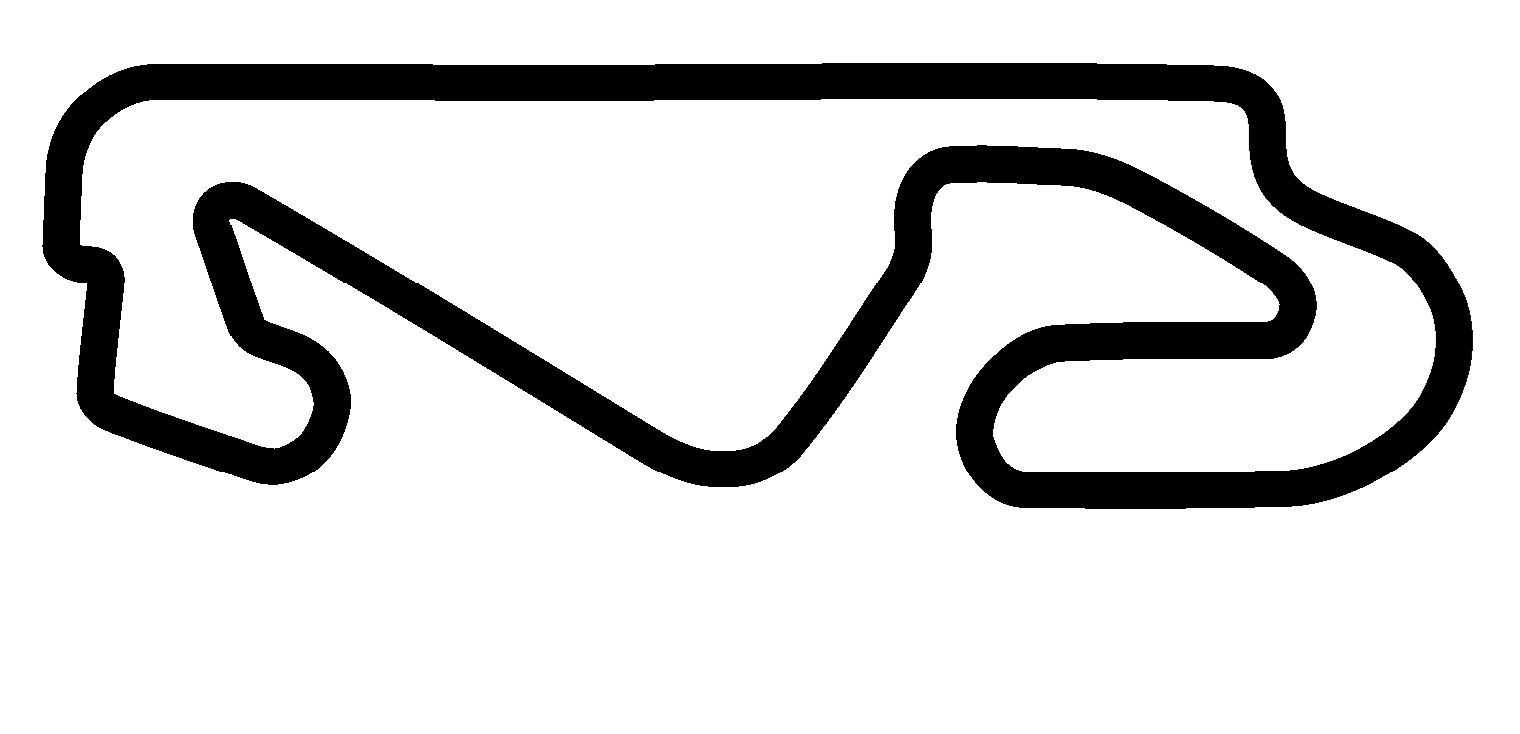}
    \includegraphics[width=0.4\linewidth]{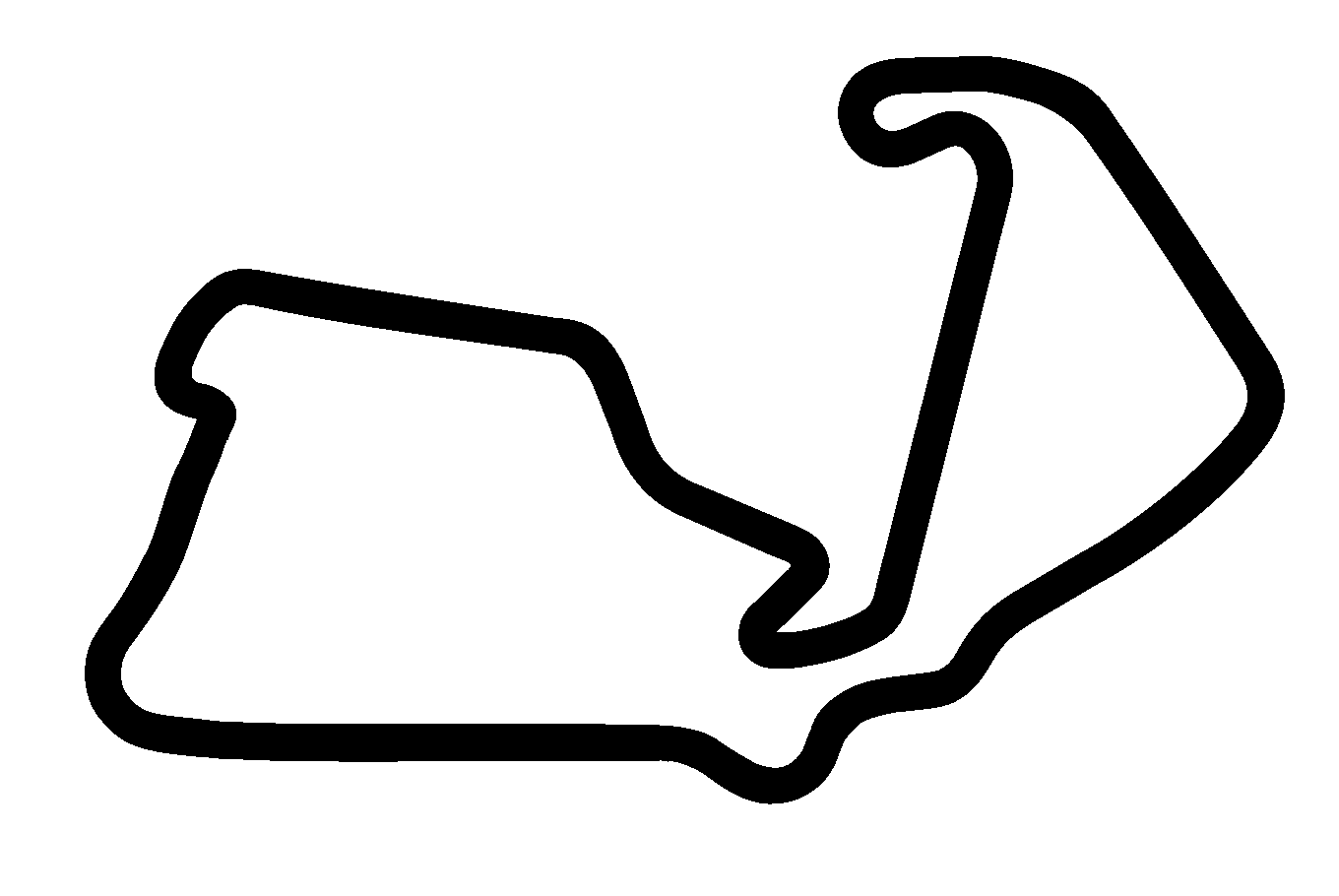}
    \caption{Track layouts of the ESP (left) and GBR (right).}
    \label{fig:maps}
\end{figure}

Agents are trained for 100k steps before being evaluated. Evaluation is conducted on the same track as training, with all experiments being repeated for two track layouts; ESP and GBR. The layout of these tracks can be seen in Fig. \ref{fig:maps}. All results are also averaged over five random seeds.

\subsection{Metrics}

In order to understand the behavior of each algorithm, we compare the performance of each algorithm using racing-related metrics. The metrics we used are as follows:

\textbf{Track Progress (PG):} This metric indicates the largest proportion of the track an agent has completed in an episode. A track progress of \(1.0\) indicates the trainable agent completed a single lap of the track, whereas a track progress of \(0.5\) indicates the agent completed half of a lap of the track before the episode ended.  

    
\textbf{Overtakes (OT):} This metric indicates the average number of overtakes performed by the trainable agent in each episode, based on track progress. For each overtake of an adversary, we increment an overtake score by one. For each time an adversary overtakes the trainable agent, we decrement the score by one. The score begins at zero.
    
\textbf{Agent-to-Agent Collisions (A2A):} This metric indicates the total amount of episodes that ended in agent-to-agent collisions between the trainable agent and an adversary. A score of \(50\) would indicate that all evaluation episodes ended with agent-to-agent collisions. 
    

%% file: results.tex
\section{Results} \label{sec:Results}

When presenting the results of our experiments, we will use the term ``in-distribution'' when referring to episodes in which both values of the context vector are within the range of context values seen during training, and the term ``out-of-distribution'' when referring to episodes in which one or both values of the context vector are outside this range. 


\subsection{Performance Benchmarking}\label{sec:performance-benchmarking}
As a preliminary result, we present Table \ref{tab:sac}, in which we observe that the SAC achieves less than 25\% track progress in both tracks, compared to approximately 50\% completed by the DreamerV3 agent. This render SAC as a nonviable method for head-to-head autonomous racing in this environment and we exclude the algorithm from further discussion. Table \ref{tab:means} reports the mean performance of the DreamerV3, cRSSM, and cMask algorithms on the ESP and GBR tracks.

The DreamerV3 agent performs a high number of overtakes in our experiments. The agent achieves the highest rate of overtakes against out-of-distribution adversaries in three experiment configurations, and the highest rate of overtakes against in-distribution adversaries in two experiment configurations. However, the agent never achieves the lowest number of agent-to-agent collisions. These results may suggest that the DreamerV3 algorithm results in a more aggressive, yet unsafe, policy when applied to autonomous racing. 

In contrast to this, the cRSSM algorithm appears to learn a more conservative driving policy. The cRSSM agent achieves the lowest rate of agent-to-agent collisions against all adversaries in two of the four experiment configurations. Against out-of-distribution adversaries in particular, the cRSSM agent collides with adversaries the least in three experiments. However, the agent achieves the highest rate of overtakes in no experiments, suggesting that the cRSSM algorithm results in a safe, yet conservative, driving policy. 

Our proposed cMask agent appears to learn a blend of aggressive and conservative driving styles. This is demonstrated by cMask agent achieving the best performance across all metrics in two experiments when adversary behavior is within the range seen during training. The agents performance drops when observing out-of-distribution dynamics, however the agent still achieves a lower rate of agent-to-agent collisions relative to DreamerV3 in three of the four experiment configurations. 

Another observation from our results is that the lowest rate of agent-to-agent collisions is achieved only by context-aware agents in each experiment configuration. This suggests that the use of context-aware MBRL algorithms results in policies that are safer in agent-to-agent interactions. 


\begin{table}
    \centering
    \normalsize
        \vspace{5pt}
        \begin{tabular*}{\linewidth}{@{\extracolsep{\fill}} l | c c | c c}
             & \multicolumn{2}{c}{\textbf{In-Distribution}} & \multicolumn{2}{c}{\textbf{Out-of-Distribution}} \\
             &  ESP & GBR &  ESP & GBR \\
            \hline
           SAC vs One & 0.2045 & 0.2357 & 0.2103 & 0.2038  \\
            SAC vs Three & 0.1857 & 0.2326 & 0.1756 & 0.1999\\
            \hline
            DRM vs One & 0.5515 & 0.5057 & 0.5255 & 0.4446  \\
            DRM vs Three & 0.5479 & 0.6243 & 0.4416 & 0.5309  \\
        \end{tabular*}
    \captionsetup{width=\linewidth}
    \caption{Track progress comparison between SAC and DreamerV3 (DRM). }
    \label{tab:sac}
\end{table}

\begin{table}
    \centering
    \small
        \vspace{5pt}
        \begin{subtable}[h]{\columnwidth}          
        \begin{tabular*}{\linewidth}{@{\extracolsep{\fill}} c l c c c c c}
            \toprule
            & & OOD? & PG [\(\uparrow\)] & OT [\(\uparrow\)] & A2A [\(\downarrow\)]\\
            \midrule
            \multirow{6}[6]{*}{\begin{turn}{90} \thead{\hspace{2mm} \footnotesize \textbf{Single} \\ \hspace{2mm} \footnotesize \textbf{Adversary}} \end{turn}}
            & DreamerV3 & $\nobox$ & 0.5515 & \textbf{0.6231} & 17.1111 \\
            & cRSSM & $\nobox$ & \textbf{0.5841} & 0.4867 &\textbf{ 9.3778} \\
            & cMask & $\nobox$ & 0.4300 & 0.2120 & 26.6444 \\
            \cline{2-6}            
            & DreamerV3 & $\yesbox$ & \textbf{0.5255} & \textbf{0.4096} & 18.5850 \\
            & cRSSM & $\yesbox$ & 0.4888 & 0.3484 & \textbf{14.6550} \\
            & cMask & $\yesbox$ & 0.4175 & 0.2113 & 23.4100\\
            \midrule
            \multirow{6}[6]{*}{\begin{turn}{90} \thead{\hspace{2mm} \footnotesize \textbf{Three} \\ \hspace{2mm} \footnotesize \textbf{Adversaries}} \end{turn}}
            & DreamerV3 & $\nobox$ &  0.5479 & 0.0667 & 18.4667 \\
            & cRSSM & $\nobox$ & 0.5664 & 0.0658 & 20.2889 \\
            & cMask & $\nobox$ & \textbf{0.6331} & \textbf{0.0867} & \textbf{14.6667} \\
            \cline{2-6}            
            & DreamerV3 & $\yesbox$ & 0.4416 & \textbf{0.2038} & 19.2650 \\
            & cRSSM & $\yesbox$ & 0.4732 & 0.0538 & 20.3500 \\
            & cMask & $\yesbox$ & \textbf{0.5335} & 0.0759 & \textbf{18.4063}\\
            \bottomrule
        \end{tabular*}
        \caption{ESP Track}
    \end{subtable}

    \vspace{2mm}

    \begin{subtable}[h]{\columnwidth}
        \begin{tabular*}{\linewidth}{@{\extracolsep{\fill}} c l c c c c c}
            \toprule
            & & OOD? & PG [\(\uparrow\)] & OT [\(\uparrow\)] & A2A [\(\downarrow\)] \\
            \midrule
            \multirow{6}[6]{*}{\begin{turn}{90} \thead{\hspace{2mm} \footnotesize \textbf{Single} \\ \hspace{2mm} \footnotesize \textbf{Adversary}} \end{turn}}
            & DreamerV3 & $\nobox$ & 0.5057 & 0.0080 & 10.1111\\
            & cRSSM & $\nobox$ &  0.4740 & 0.0342 &  10.6444  \\
            & cMask & $\nobox$ & \textbf{0.5993} &\textbf{ 0.0474} & \textbf{9.6222} \\
            \cline{2-6}            
            & DreamerV3 & $\yesbox$ & 0.4446 & \textbf{0.4096} & 14.2650\\
            & cRSSM & $\yesbox$ &  0.4436 & 0.0542 &  \textbf{12.3750}  \\
            & cMask & $\yesbox$ & \textbf{0.5463} & 0.0356 & 13.2700\\
            \midrule
            \multirow{6}[6]{*}{\begin{turn}{90} \thead{\hspace{2mm} \footnotesize \textbf{Three} \\ \hspace{2mm} \footnotesize \textbf{Adversaries}} \end{turn}}
            & DreamerV3 & $\nobox$ & 0.6243 & \textbf{0.0529} & 15.0444 \\
            & cRSSM & $\nobox$ & \textbf{0.6419} & 0.0036 &  \textbf{3.6888}  \\
            & cMask & $\nobox$ & 0.5395 & 0.0387 & 12.8000 \\
            \cline{2-6}            
            & DreamerV3 & $\yesbox$ &  0.5309 & 0.1217 & 19.1349 \\
            & cRSSM & $\yesbox$ &  \textbf{0.5481} & \textbf{0.2820} &  \textbf{10.5350}  \\
            & cMask & $\yesbox$ & 0.4924 & 0.0395 & 14.8900\\
            \bottomrule
        \end{tabular*}
    \caption{GBR Track}
    \end{subtable}
    
    \captionsetup{width=\linewidth}
    \caption{Mean performance metrics. Rows marked ``\textbf{OOD?}'' indicate measurements recorded against adversaries with out-of-distribution driving behaviors.}
    \label{tab:means}
\end{table}

\subsection{Generalization Benchmarking}

In order to quantify how well each algorithm generalizes to unseen adversarial driving behaviors, we have calculated the percentage relative change between each algorithms mean in-distribution and out-of-distribution performances. The intuition behind this choice is that an agent with poor generalization will experience a drop in performance when observing unseen dynamics, whereas a generalizable algorithm will see no such drop or experience a positive change. We have omitted the use of the overtaking metric in this comparison as out-of-distribution contexts may change the underlying difficulty of performing an overtake, e.g., the adversary driving faster than the trainable agent. Therefore, the overtaking metric may not be a useful indicator of generalization capabilities. Table \ref{tab:rel} displays percentage change in performance of each algorithm when evaluated on out-of-distribution data for the ESP and GBR tracks.

We observe that our proposed cMask algorithm has the best relative performance in two of the four experiment configurations for both the progress metric and the agent-to-agent collision metric. Furthermore, if we exclude the DreamerV3 algorithm, it achieves the best performance in the progress metric for an additional configuration. These data suggest that the cMask agent has strong generalization capabilities relative to the other MBRL approaches, even context-aware approaches. One may also observe that the DreamerV3 agent only achieves the best relative performance in one metric for a single configuration. From this one may infer that the context-free MBRL method, DreamerV3, generalizes poorly to unseen data relative to the context-aware MBRL methods.

\begin{table}
    \normalsize
    \centering
    \vspace{5pt}
    \begin{subtable}[h]{\columnwidth}
    \centering
        \begin{tabular}[t]{@{\extracolsep{\fill}} c l c c}
            \toprule
            & & \(\Delta_{PG}\%\) [\(\uparrow\)] & \(\Delta_{A2A}\%\) [\(\downarrow\)] \\
            \midrule
            \multirow{3}[3]{*}{\begin{turn}{90} \thead{\hspace{2mm} \small \textbf{Single} \\ \hspace{2mm} \small \textbf{Adv}} \end{turn}}
            & DreamerV3 & \textbf{-0.6527} & +8.6137 \\
            & cRSSM & -16.3157 & +53.2733 \\
            & cMask & -2.9070 & \textbf{-12.1391} \\
            \midrule
            \multirow{3}[3]{*}{\begin{turn}{90} \thead{\hspace{2mm} \small \textbf{Three} \\ \hspace{2mm} \small \textbf{Advs}} \end{turn}}
            & DreamerV3 & -19.4014 & +4.3229 \\
            & cRSSM & -16.4548 &\textbf{ +0.3011} \\
            & cMask & \textbf{-15.7321} & +25.4972 \\
            \bottomrule
        \end{tabular}
        \caption{ESP Track}
    \end{subtable}

    \vspace{2mm}
        
    \begin{subtable}[h]{\columnwidth}
    \centering
        \begin{tabular}[t]{@{\extracolsep{\fill}} c l c c}
            \toprule
            & & \(\Delta_{PG}\%\) [\(\uparrow\)] & \(\Delta_{A2A}\%\) [\(\downarrow\)] \\
            \midrule
            \multirow{3}[3]{*}{\begin{turn}{90} \thead{\hspace{2mm} \small \textbf{Single} \\ \hspace{2mm} \small \textbf{Adv}} \end{turn}}
            & DreamerV3 & -12.0823 & +41.0826 \\
            & cRSSM & \textbf{-6.4135} & \textbf{+16.2583} \\
            & cMask & -8.8437 & +37.9102 \\
            \midrule
            \multirow{3}[3]{*}{\begin{turn}{90} \thead{\hspace{2mm} \small \textbf{Three} \\ \hspace{2mm} \small \textbf{Advs}} \end{turn}}
            & DreamerV3 & -14.9608 & +27.1895 \\
            & cRSSM & -14.6129 & +185.5942 \\
            & cMask & \textbf{-8.7303} & \textbf{+16.3281} \\
            \bottomrule
        \end{tabular}
        \caption{GBR Track}
    \end{subtable}
    
    \captionsetup{width=\linewidth}
    \caption{Percentage difference between in-distribution performance and out-of-distribution performance.}
    \label{tab:rel}
\end{table}

%% file: conclusion.tex
\section{Conclusion}
We presented our work on the performance and generalization capabilities of MBRL algorithms in the simulated autonomous racing environment, Roboracer. We frame the task of head-to-head racing in this environment as a contextual RL problem by parameterizing adversary driving behavior by the context of the episode. This framework allowed us to evaluate RL agents subject to out-of-distribution transition and reward dynamics. In this setting, we trained and evaluated a set of MBRL algorithms, including a novel context-aware approach introduced in this paper, cMask. 

Our results suggest that context-aware MBRL agents display safer behaviors relative to context-free MBRL agents when interacting with other agents in the environment. We also demonstrate that context-aware approaches experience a lesser susceptibility to reductions in performance when observing out-of-distribution transition dynamics. Finally, we suggest that the context masking technique presented in this work may enable better in-distribution performance for context-aware MBRL methods for autonomous racing.

The limitations of this work include assuming that the context is fixed and observable, and that the behavior of all adversarial agents can be fully parameterized by it, which restricts applicability to real-world robotics. Future work may address these limitations by developing methods to predict more flexible representations of context.


%% file: acknowledgment.tex
\section*{Acknowledgment}
This publication has in part emanated from research supported by Taighde Éireann – Research Ireland under Frontiers for the Future Grant No. 21/FFP-A/8957. For the purpose of Open Access, the author has applied a CC BY public copyright license to any Author Accepted Manuscript version arising from this submission.

%% file: bibliography.tex
\bibliographystyle{IEEEtran}
\bibliography{ref}